\documentclass[letterpaper, 10 pt, conference]{ieeeconf}  

\IEEEoverridecommandlockouts                              

\usepackage{cite}
\usepackage{amsmath,amssymb,amsfonts}
\usepackage{algorithmic}
\usepackage{graphicx}
\usepackage{textcomp}
\usepackage{xcolor}
\usepackage{todonotes}

\usepackage[pdftex,unicode, 
colorlinks=true,
linkcolor = black]{hyperref}

\def\BibTeX{{\rm B\kern-.05em{\sc i\kern-.025em b}\kern-.08em
    T\kern-.1667em\lower.7ex\hbox{E}\kern-.125emX}}

\usepackage[
singlelinecheck=false 
]{caption}

\title{\LARGE \bf
Raindrops on Windshield: Dataset and Lightweight Gradient-Based Detection Algorithm*
}

\author{Vera Soboleva$^{1,2,3}$ and Oleg Shipitko$^{1,2}$
\thanks{*This work was supported by Evocargo LLC}
\thanks{$^{1}$Vera Soboleva and Oleg Shipitko are with Evocargo LLC, Moscow, Russia
        {\tt\small vera.burdina@evocargo.com}, {\tt\small oleg.shipitko@evocargo.com}}%
\thanks{$^{2}$Vera Soboleva and Oleg Shipitko are with the Vision Systems Laboratory, Institute for Information Transmission Problems, Moscow, Russia}
\thanks{$^{3}$Vera Soboleva is with the Moscow Institute of Physics and Technology (National Research University), Dolgoprudny, Russia}
}

\begin{document}

\maketitle

\thispagestyle{empty}

\pagestyle{empty}

\begin{abstract}
Autonomous vehicles use cameras as one of the primary sources of information about the environment. Adverse weather conditions such as raindrops, snow, mud, and others, can lead to various image artifacts. Such artifacts significantly degrade the quality and reliability of the obtained visual data and can lead to accidents if they are not detected in time. This paper presents ongoing work on a new dataset for training and assessing vision algorithms' performance for different tasks of image artifacts detection on either camera lens or windshield. At the moment, we present a publicly available set of images containing $8190$ images, of which $3390$ contain raindrops. Images are annotated with the binary mask representing areas with raindrops. We demonstrate the applicability of the dataset in the problems of raindrops presence detection and raindrop region segmentation. To augment the data, we also propose an algorithm for data augmentation which allows the generation of synthetic raindrops on images. Apart from the dataset, we present a novel gradient-based algorithm for raindrop presence detection in a video sequence. The experimental evaluation proves that the algorithm reliably detects raindrops. Moreover, compared with the state-of-the-art cross-correlation-based algorithm \cite{Einecke2014}, the proposed algorithm showed a higher quality of raindrop presence detection and image processing speed, making it applicable for the self-check procedure of real autonomous systems. The dataset is available at \href{https://github.com/EvoCargo/RaindropsOnWindshield}{$github.com/EvoCargo/RaindropsOnWindshield$}.
\end{abstract}


\section{Introduction}
As a part of the Intelligent Transportation System, autonomous vehicles use video cameras to perceive the environment. They are used to detect objects on the road and navigate. However, adverse weather conditions can negatively affect the performance of the system. Raindrops or snow could block the camera's view and make the further movement of autonomous vehicles unsafe for both passengers and pedestrians. In this regard, it is essential that autonomous transportation systems can run self-check procedures and evaluate whether they can still perform their intended function. 

The main difficulty in image artifacts detection is that drops and other contaminants can have all kinds of shapes, shades, and structures, which significantly complicates their parametric description. Besides, artifact detection algorithms should meet high requirements on speed and computational efficiency. On the one hand, they must work in real-time, and on the other hand, they are supplemental procedures in the autonomous system functioning and should not be a significant consumer of computing resources.

\begin{figure}[t!]
    \begin{minipage}[h]{0.47\linewidth}
    \center{\includegraphics[width=1\linewidth, height=0.1\textheight]{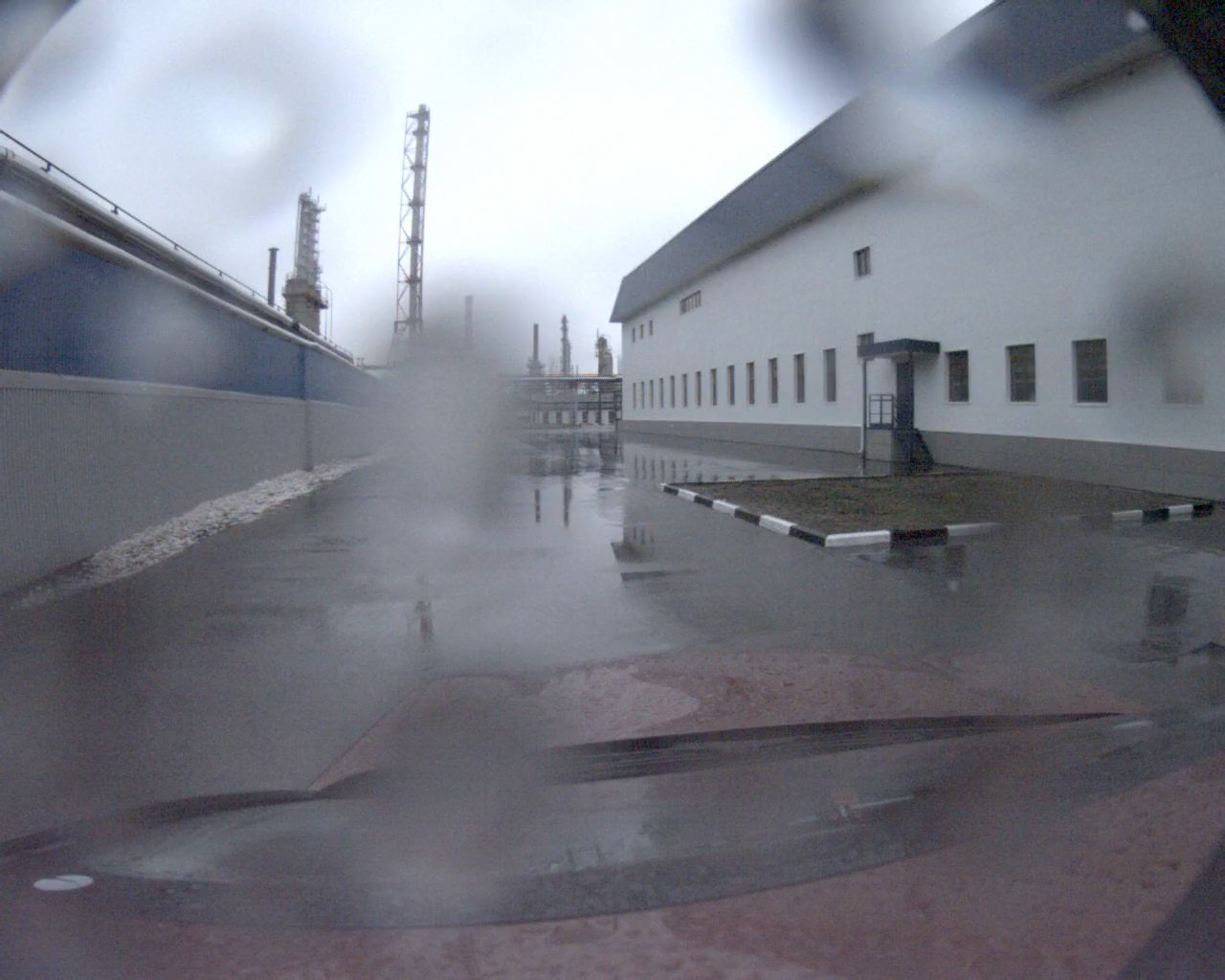}} (a) original image \\
    \end{minipage}
    \hfill
    \begin{minipage}[h]{0.47\linewidth}
    \center{\includegraphics[width=1\linewidth, height=0.1\textheight]{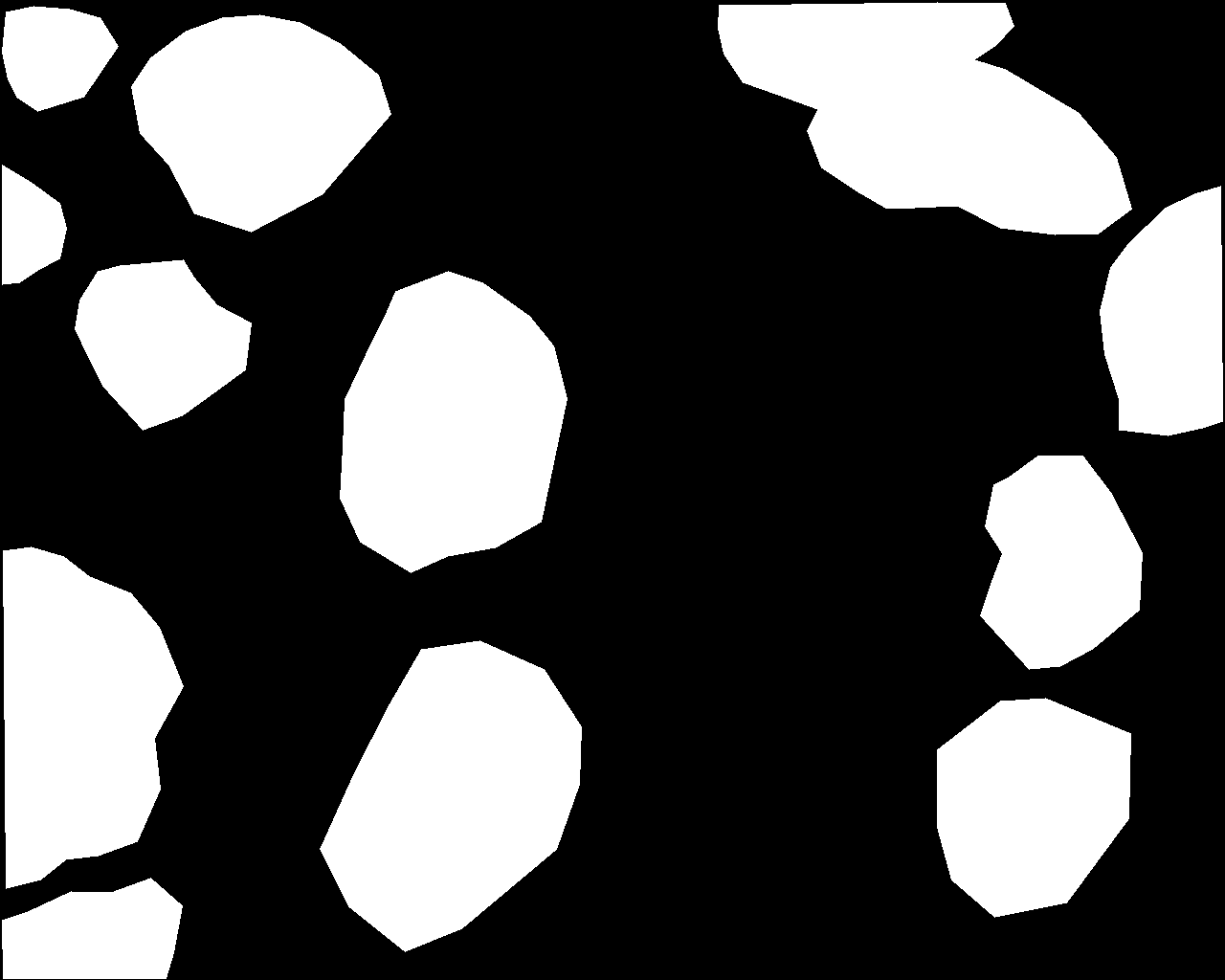}} \\(b) mask
    \end{minipage}
    \vfill
    \begin{minipage}[h]{0.47\linewidth}
    \center{\includegraphics[width=1\linewidth]{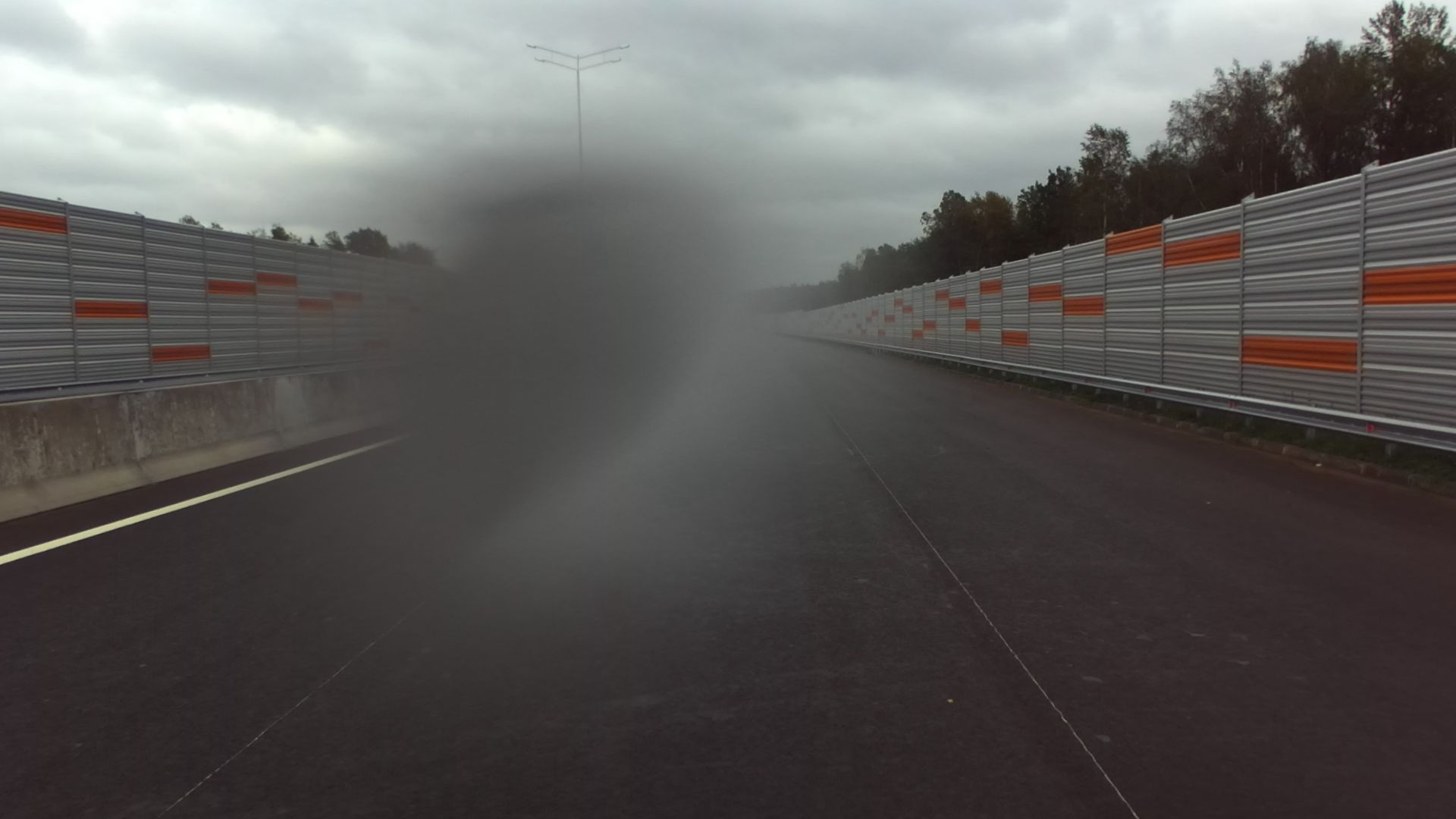}} (c) original image \\
    \end{minipage}
    \hfill
    \begin{minipage}[h]{0.47\linewidth}
    \center{\includegraphics[width=1\linewidth]{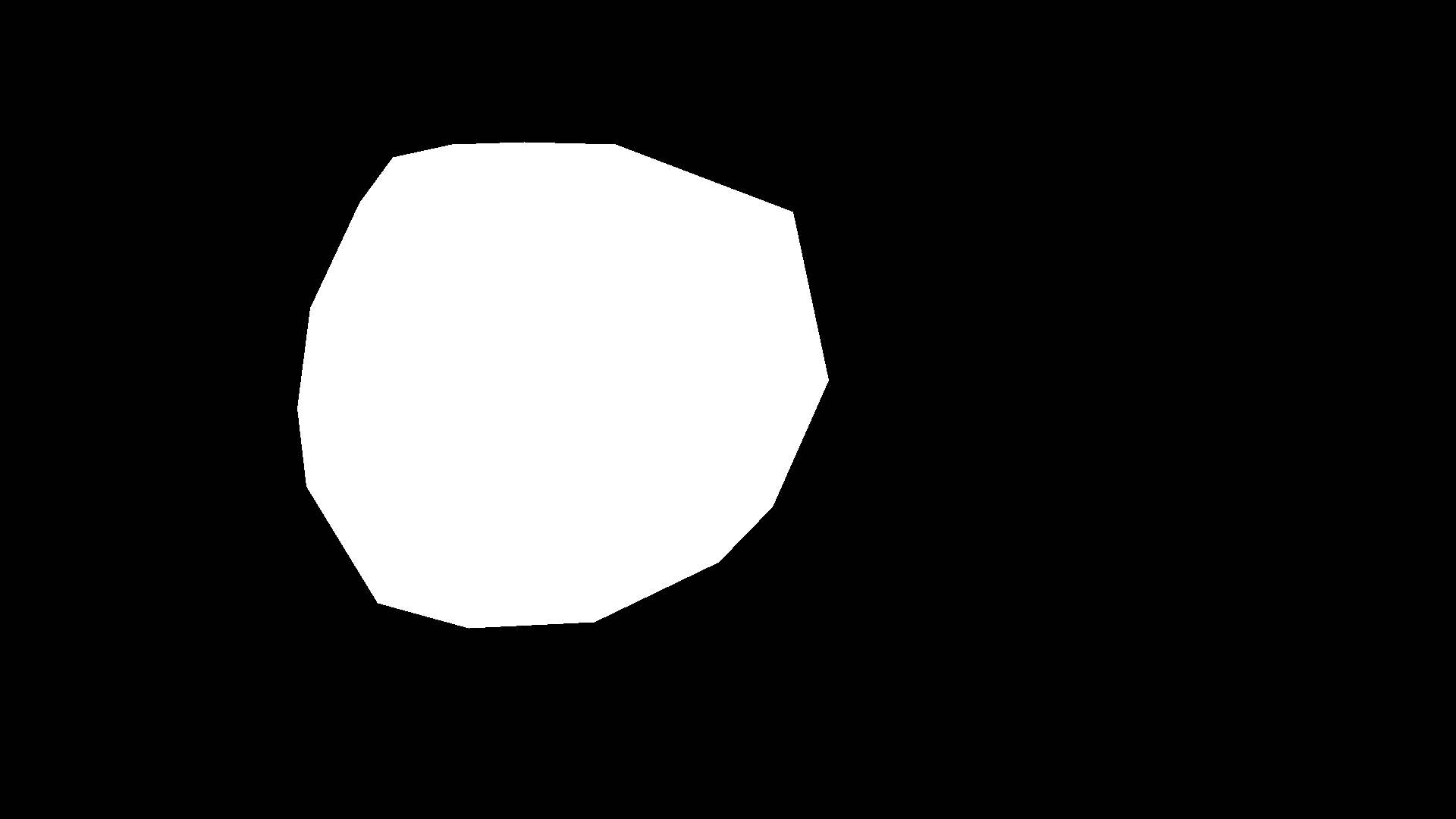}} (d) mask\\
    \end{minipage}
    \caption{Example of dataset samples labeling. White color denotes an area with an artifact.}
    \label{fig:labeling}
\end{figure}

In this paper, we present a new publicly available dataset for training and evaluating algorithms for various problems of detecting raindrops on an image. The dataset contains sequences of images captured by a camera attached to the vehicle during its movement. At the moment, it consists of $8190$ images, $3390$ of which contain raindrops. All images are annotated with the binary mask representing areas with raindrops. Annotation examples can be seen in Fig.~\ref{fig:labeling}. We demonstrate the applicability of the dataset in the problems of raindrop presence detection and raindrop segmentation. In addition to the dataset, we present an artificial raindrops generation algorithm, which allows the creation of various realistic raindrops to augment the data. 

Apart from the dataset, we propose a novel algorithm for detecting raindrops on a sequence of images. The algorithm is based on two assumptions. Firstly, as observations show, raindrops mostly remain stable in their position in the images. Secondly, the area covered by the drop becomes blurred, so there are no sharp boundaries inside the raindrop image area. Thus, the sequence of images is analyzed for the presence of the static regions without sharp edges inside. For this, a gradient map is calculated for each frame, which is then averaged over the entire sequence. The problem of detecting raindrops was considered as a problem of binary classification of image sequences. The performance of the algorithm was evaluated using ROC curve and AUC-ROC metric.
The experimental evaluation showed that the algorithm reliably detects raindrops and demonstrates better classification quality and higher image processing speed than the correlation-based algorithm~\cite{Einecke2014}. Thus, a gradient-based algorithm can be applied as a real-time procedure for autonomous visual systems self-test.

The proposed algorithm can also solve the raindrop segmentation problem. However, modern neural network approaches will indeed outperform it. In this paper, we trained the original U-Net architecture's neural network~\cite{UNet2015} to compare our algorithm with it and demonstrate the dataset's applicability for this task. We evaluated the segmentation quality via the Dice coefficient calculated over raindrops pixels, intersection over union (IoU), and pixel-wise accuracy. According to the experiment results, U-Net showed high metrics values on the test samples, which means that the proposed dataset is suitable for training neural networks.

\section{Related work} 
Most of the existing weather conditions datasets are not suitable for solving the problem of vehicle camera lens pollution detection and do not provide the ability to work with a sequence of images since they contain independent frames. For instance, Multi-class Weather Image~\cite{Zhang2015} and Image2Weather~\cite{Chu2017} datasets are only intended for classifying images by weather but not for determining the presence of artifacts on the camera lens.
On the other hand, there are visual navigation aimed datasets such as the Oxford RobotCar dataset~\cite{RobotCar2017} or KITTY~\cite{geiger2013vision}, containing image sequences with various weather conditions. However, unlike our dataset, they are not suitable for the raindrop segmentation problem since it requires additional ground-truth mask labeling.

There are many works related to image artifact detection. Some of them use a sequence of images, while others perform detection on a single image. Algorithms from~\cite {Einecke2014, You2015, Cord2011, Nashashibi2010} are based on the assumption that most artifacts are temporally stable in their position in the image. Therefore they use a sequence of frames for detection.  

In \cite {Einecke2014} a pixel-wise normalized cross-correlation between pairs of images is calculated to analyze a sequence of images for static regions. The idea is that the corresponding pixels of two consecutive frames belonging to static objects will show high correlation values. At the same time, the moving parts of the scene will be weakly correlated. The main disadvantage of this approach is that, in linear motion, distant objects remain almost stable in their positions in the image over time and give high correlation values, leading to false artifacts detections. Also, if the artifacts are very opaque and the intensity variance inside the image area corresponding to them is close to zero, the algorithm becomes computationally unstable. Moreover, if the artifacts, on the contrary, are too transparent, then the intensity inside them changes significantly, which leads to low correlation values and, therefore, false-negative detection results.

The authors of~\cite {You2015} also assume that intensity change in raindrop area is less than in area without it. Therefore, the pixel intensities' spatial and temporal derivatives are analyzed to detect artifacts. For this, the optical flow and the pixel-wise change in intensity over time are calculated. The major drawback of this method is that the algorithm can not work in real-time because computing the optical flow of all pixels has high computational complexity. If only pixel intensities' temporal derivatives are used, the algorithm becomes too sensitive to lighting conditions changes, making it unsuitable for outdoor use. It should be noted that in~\cite {You2015}, as well as in this work, the artificial raindrops generation method is used to expand the dataset.

In ~\cite {Cord2011} a gradient map averaged over several frames is investigated, to which nonlinear filtering is applied. The main difference between the algorithm proposed in this work and ~\cite {Cord2011} is that the algorithm from ~\cite {Cord2011} solves the problem of detecting tiny drops similar to point noise in the image. Gradient maps are also used in ~\cite {Nashashibi2010}, where the appearance of a drop in the image is detected by changing the intensities between two sequential frames. Authors use an image gradient to check if the candidate region's boundaries have either disappeared or become more blurred over time.  The assumption that the image area overlapped by the drop becomes blurred and does not have sharp boundaries inside is also used in our algorithm.

There are also single image artifact detection algorithms. In ~\cite {Roser2009}, a raindrop pattern is artificially created based on a photometric raindrop model. Detection is then performed by comparing the generated raindrops with the original image regions. The main disadvantage of this method is that it detects only drops of a specific shape.

The authors of \cite{Akkala2016} propose to detect raindrops by detecting blurred regions. They prove that blurred image areas, which are raindrops, correspond to a lower Kurtosis coefficient than clear ones. Also, they use  Discrete Wavelet Transform to validate the top of the image (the sky) and Singular Value Decomposition to validate the bottom (the road). The downside is that the algorithm can not work with trees at the top of the image and too dense drops and dirt.

There is another group of approaches that solve the problem with neural networks. In~\cite {Ivanov2019} simplified modifications of VGG, ResNet, and InceptionV3 architectures are used for image classification. The time required to classify one image (540x300 in size) by the degree and type of pollution is $8-12$ ms with more than 96\% accuracy. The main disadvantage of this approach is the significant consumption of computing power by the model. The same disadvantage is possessed by the method proposed in~\cite {Hu2018}, where VGG-16 is used to detect raindrops. To solve this problem, the authors of ~\cite{Bae2019} demonstrate a modified VGG-16. They significantly reduce the number of convolutional layers, which allows them to reduce the number of model parameters by about $250$ times.

In addition, there are several works devoted to removing raindrops from images~\cite{Qian2018, Gu2019, Hirohashi2020}. However, using such algorithms in autonomous transportation systems can be dangerous since it is never fully known what the algorithm will recreate in place of the former drop.

\section{Dataset Description}
The images for the dataset were captured by a camera attached to the vehicle during its movement. The vehicle's movement took place in urban areas and highways, making the dataset ideal for training and assessing vision algorithms for autonomous vehicle camera lens pollution detection. 

The dataset represents sequences of video frames containing $8190$ images, of which $3390$ contain raindrops. Images were labeled by outlining artifacts with polygons. Labeling results are stored in JSON format. Besides, binary masks were generated from this markup, which are also presented in the dataset for convenience(Fig.~\ref{fig:labeling}). White color denotes an artifact area.

Most weather-related artifacts on a windshield or a camera lens have the property of being almost stable in position over time. Therefore, there is no need to label each image in the sequence separately. It is sufficient to manually label the artifact on the image when it first appears and then copy labeling for all subsequent frames where the artifact is in the same position. In this work, every $50-100$-th image was manually labeled from the moment the artifact first appeared.

\section{Artificial raindrops generation algorithm}
\begin{center}
    \begin{figure}[t]
    \centering
    \includegraphics[width = 1\linewidth]{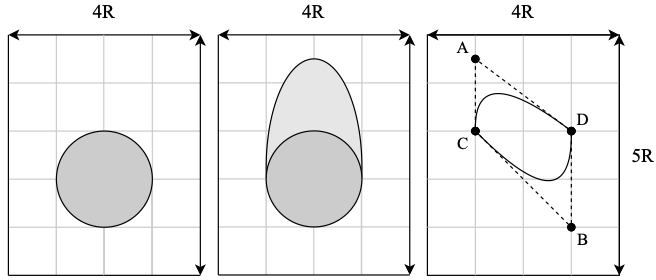}
    \caption{ Geometry of artificial drop with different shapes. From left to right: circle, egg, Bezier curves.}
    \label{fig: drop geometry}
    \end{figure}
\end{center}

\begin{center}
    \begin{figure}[t]
    \centering
    \includegraphics[width = 1\linewidth]{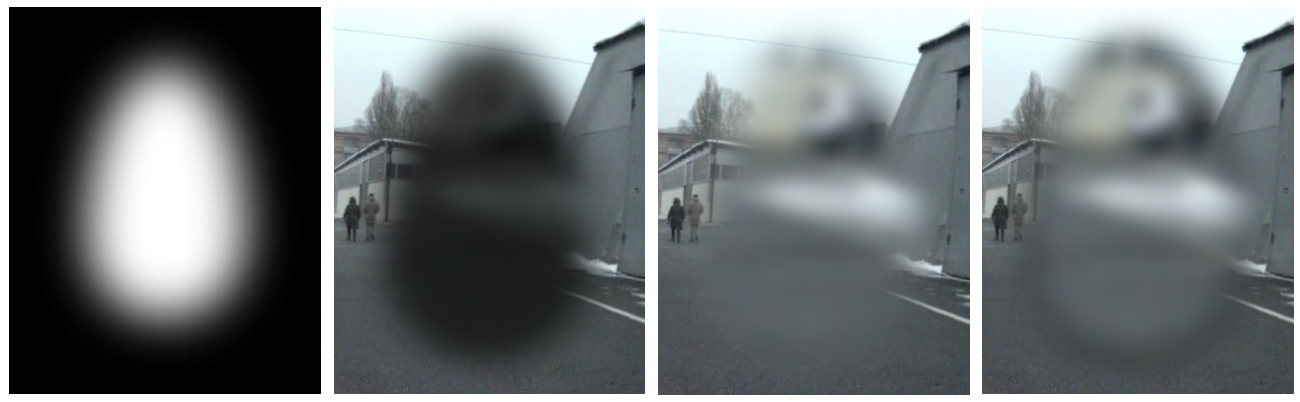} \\ (a) \ \ \ \ \ \ \ \ \ \ \ \ \ \ (b)  \ \ \ \ \ \ \ \ \ \ \ \ \ \ (c)  \ \ \ \ \ \ \ \ \ \ \ \ \ \ (d)
    \caption{Example of generation of a drop with egg shape. (a) - $AlphaMap$; (b) - result of pasting $P^{'}$ into original image, using $AlphaMap$ as a transparency mask; (c) - result of pasting $P^{'}$ into original image, using $AlphaMap$ as a transperency mask, without pasting $P$ before; (d) - result of of sequential pasting of images $P^{'}$ and $P$ into the original image using $AlphaMap$ as a transparency mask. }
    \label{fig: drop formation}
    \end{figure}
\end{center}

Collecting images with a variety of raindrops is a challenging and time-consuming task. Thus, an image raindrop simulator would greatly simplify the process and provide data augmentation to efficiently train and evaluate different raindrop detection algorithms. In this paper, we propose an algorithm that can generate diverse and realistic artificial drops in images.
The following steps describe the algorithm for generating one drop on one image $A$ of size $(W, H)$.
\begin{itemize}
\item A raindrop is assigned a random radius $R\in[R_{min}, R_{max}]$, coordinates of the drop center $ (x, y) \in (0..W, 0..H) $  and a shape: 0 -- circle , 1 -- egg, 2 -- a combination of two Bezier curves (Fig.~\ref{fig: drop geometry}). The egg shape is created by a combination of a circle and a semi-ellipse.
\item An $AlphaMap$ is created -- an empty (all zeros) grayscale image of size $(5R, 4R)$. A figure corresponding to the drop shape is drawn in the center of the image. Depending on the selected in advance brightness (0-255), the generated raindrop will be either more transparent or opaque. Next, a Gaussian blur is applied to $AlphaMap$ to make the resulting drop more realistically blurred.
\item Then, an image $P$ of size $(5R, 4R)$ is created by cropping a rectangle corresponding to the drop center from image $A$. Gaussian blur and barrel-shaped radial distortion (fish-eye effect) are applied to it.
\item An image $P^{'}$ is created by reducing the brightness of the image $P$ with a factor of $0.3$ (factor of $1.0$ corresponds to the original image, $0.0$ - to completely black). After which $P^{'}$ is inserted at the appropriate place in the original image $A$ using $AlphaMap$ as a transparency map. Where the $AlphaMap$ is $255$, the given image $P^{'}$ is copied as-is. Where the $AlphaMap$ is 0, the current value of $A$ is preserved. Intermediate values will mix the two images together. It is done to make the edges of the generated drop darker, as natural drops usually have this property.
\item Finally, using $AlphaMap$ as a transparency map, the $P$ image is pasted to the appropriate place on top of figure $P^{'}$ to the original image $A$.
\end{itemize}
Fig.~\ref{fig: drop formation} shows an example of egg-shaped drop generation. 

\section{Gradient-based algorithm for raindrops detection}
Gradient maps are widely used in image processing to highlight the edges of objects in the image: the edges in the image have a high gradient magnitude, while the homogeneous areas have a low gradient.

The area of the image covered by a raindrop is blurred and, as a result, does not have sharp boundaries inside, which means that the corresponding pixels will have a low gradient value and will be dark on the gradient map. Due to the raindrops positions stability in the image, the corresponding areas will remain dark for at least some sequence of images with raindrops. At the same time, the areas in which there is movement will have moving boundaries, which means that the gradient in them will be larger. The algorithm proposed in this work is based on this observation and uses the Sobel operator to compute gradient maps.

The input of the algorithm is a sequence of $N$ grayscale images. For each individual image, the horizontal and vertical gradients are calculated - $G_x^{(n)}(x, y)$ and $G_y^{(n)}(x, y)$, respectively, $n = \overline{1, N }$, where $(x, y)$ are the coordinates of the pixel in the image. They are used to build a gradient map $G^{(n)}(x, y) = \sqrt{G_x^{(n) 2}(x, y) + G_y^{(n) 2}(x, y)}$. Next, the gradient is averaged pixel-wise over the entire sequence:
$$G^{*}(x,y)= \frac{1}{N} \sum_{n=1}^N G^{(n)} (x,y).$$

\begin{figure}[t]
    \begin{minipage}[h]{0.47\linewidth}
    \center{\includegraphics[width=1\linewidth]{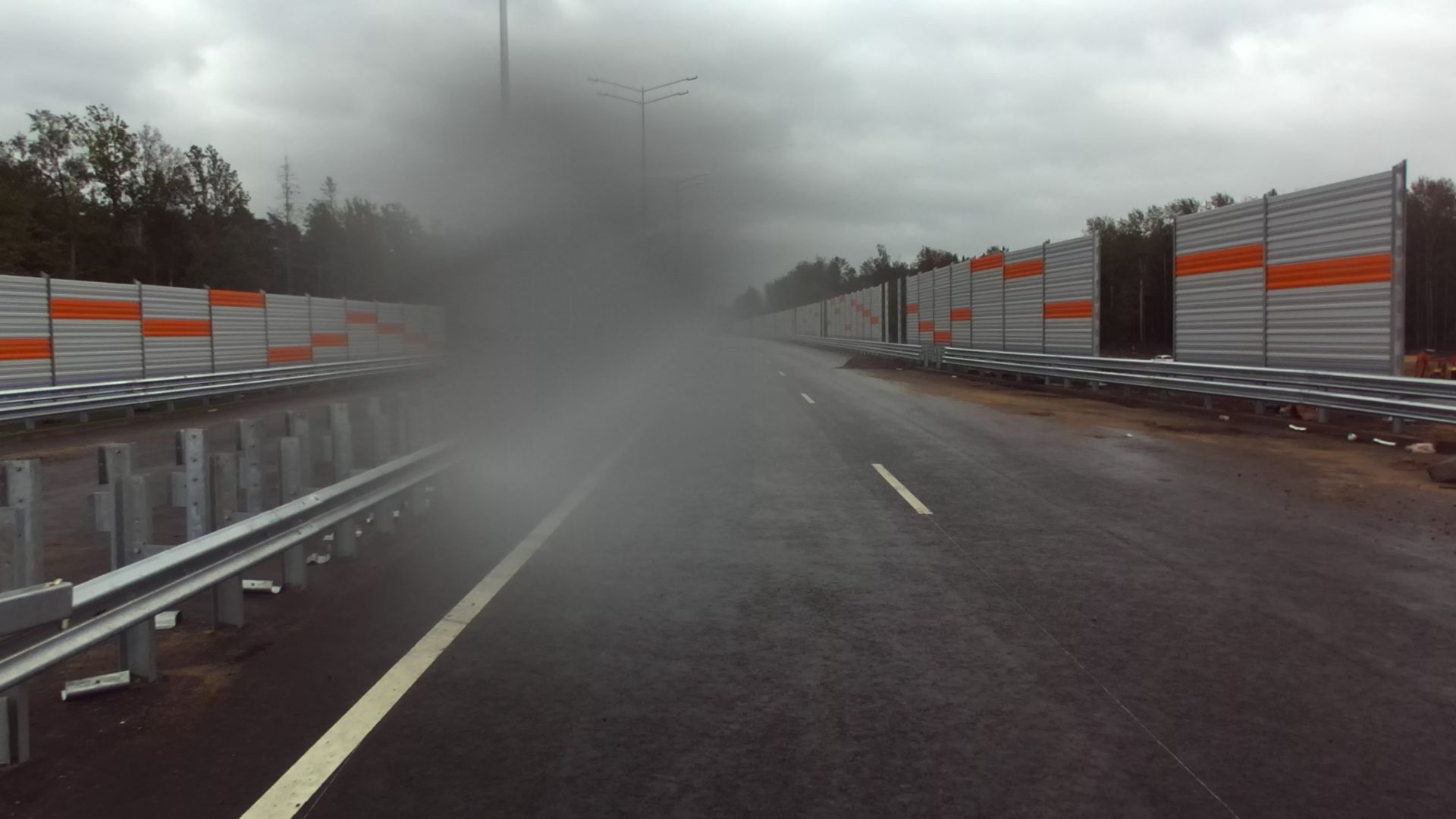}} (a) \\
    \end{minipage}
    \hfill
    \begin{minipage}[h]{0.47\linewidth}
    \center{\includegraphics[width=1\linewidth]{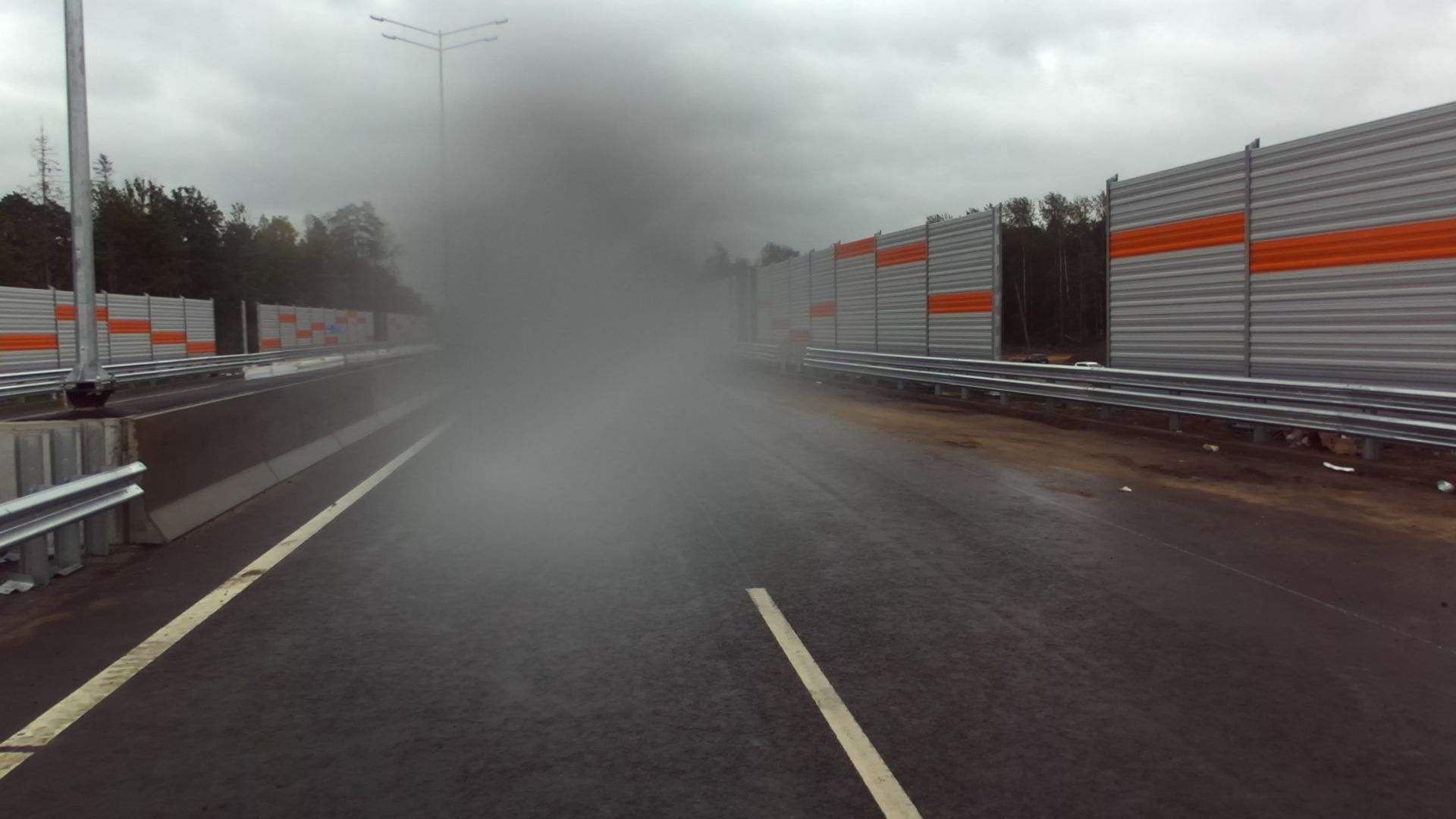}} (b) \\
    \end{minipage}
    \vfill
    \begin{minipage}[h]{0.47\linewidth}
    \center{\includegraphics[width=1\linewidth]{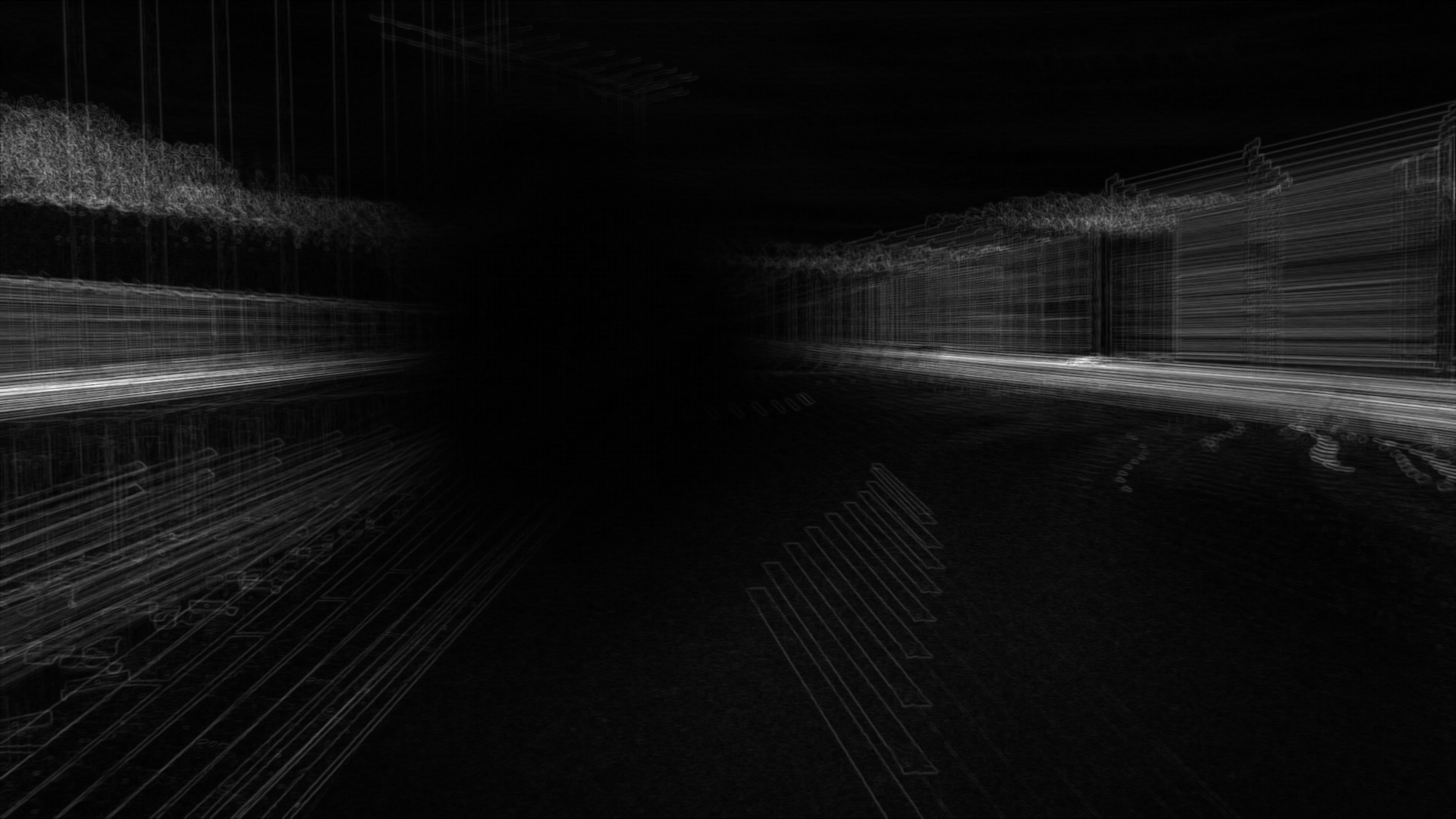}} (c) \\
    \end{minipage}
    \hfill
    \begin{minipage}[h]{0.47\linewidth}
    \center{\includegraphics[width=1\linewidth]{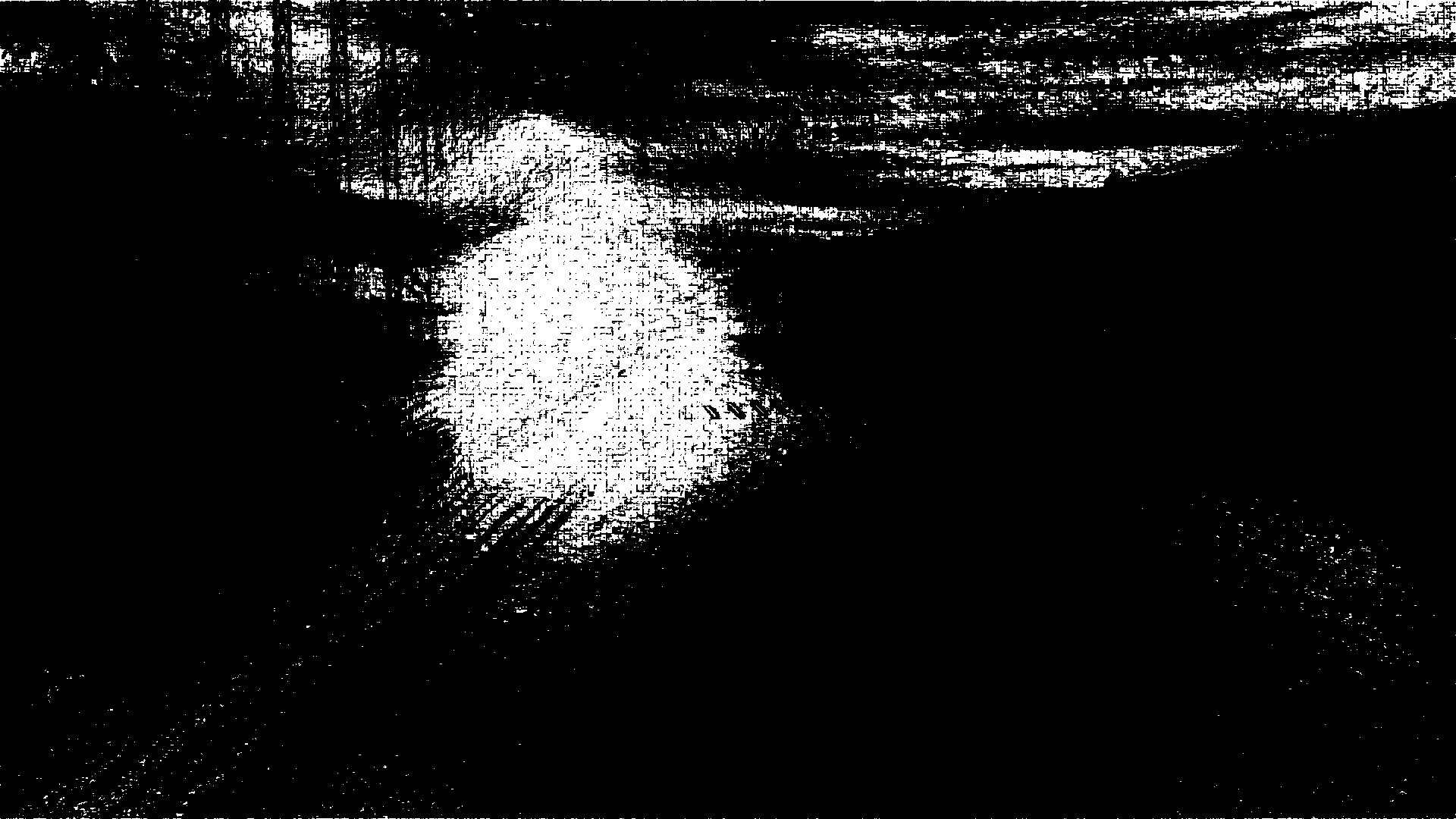}} (d) \\
    \end{minipage}
    \vfill
    \begin{minipage}[h]{0.47\linewidth}
    \center{\includegraphics[width=1\linewidth]{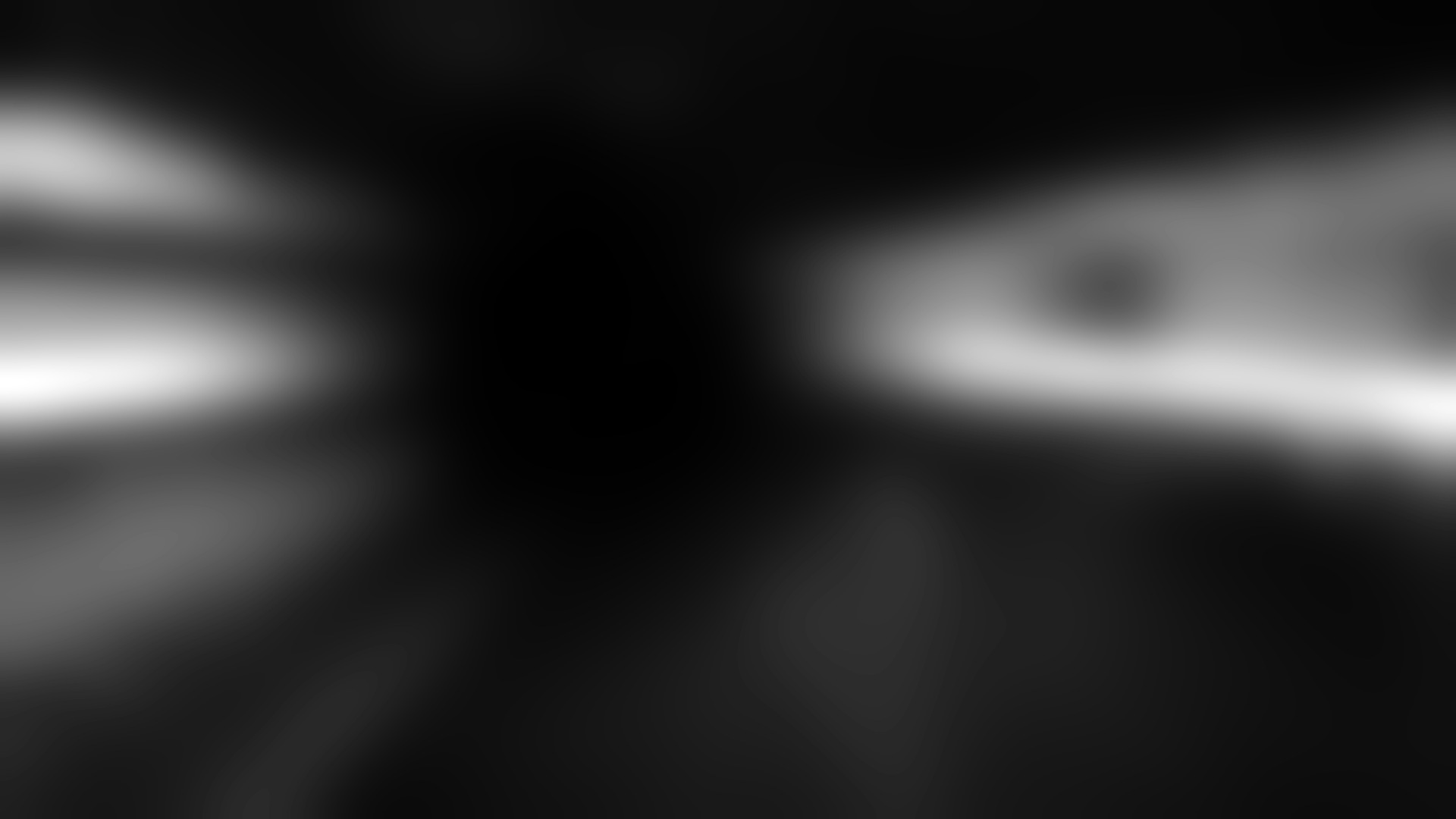}} (e) \\
    \end{minipage}
    \hfill
    \begin{minipage}[h]{0.47\linewidth}
    \center{\includegraphics[width=1\linewidth]{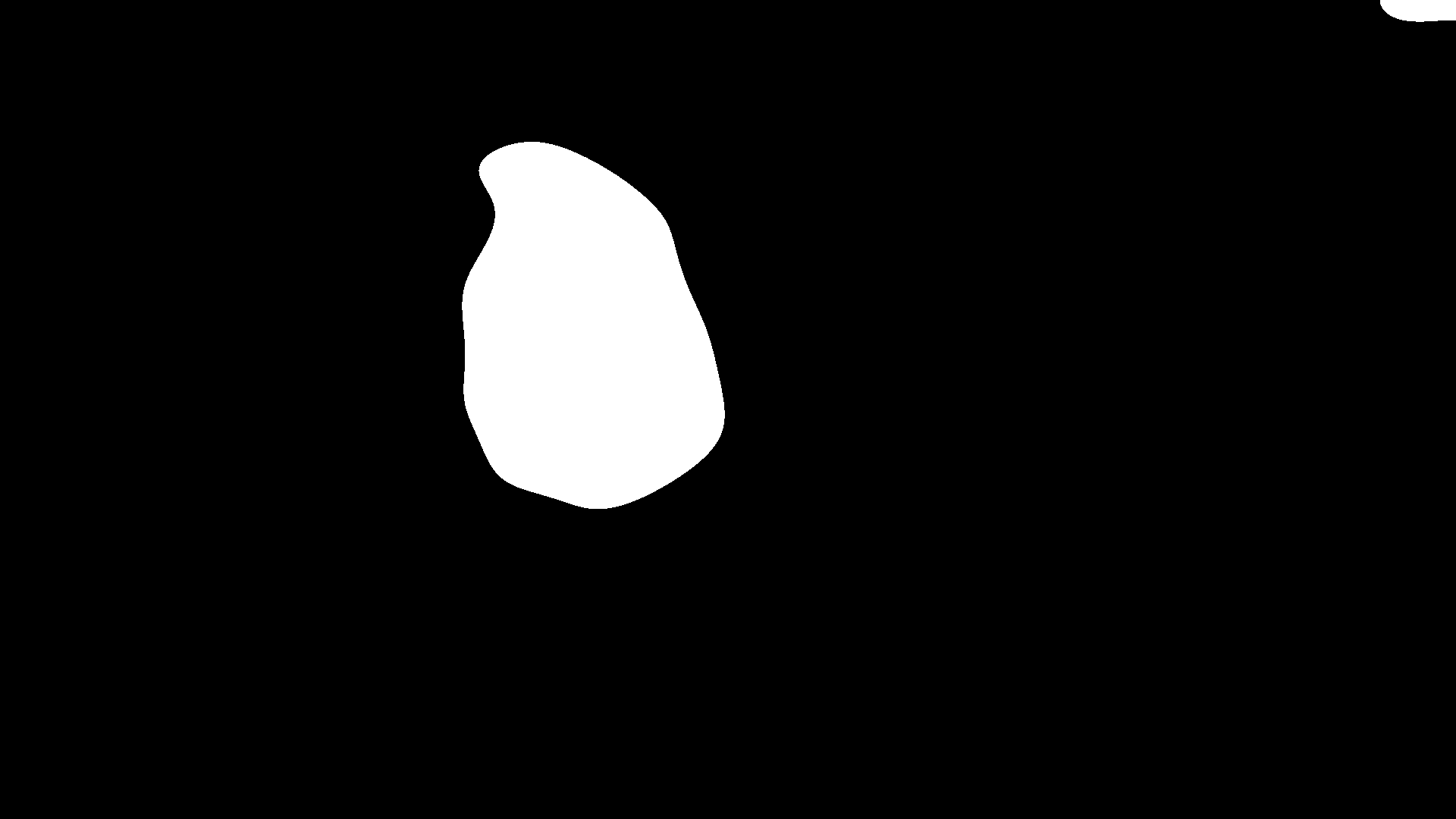}} (f) \\
    \end{minipage}
    \vfill
    \begin{minipage}[h]{0.47\linewidth}
    \center{\includegraphics[width=1\linewidth]{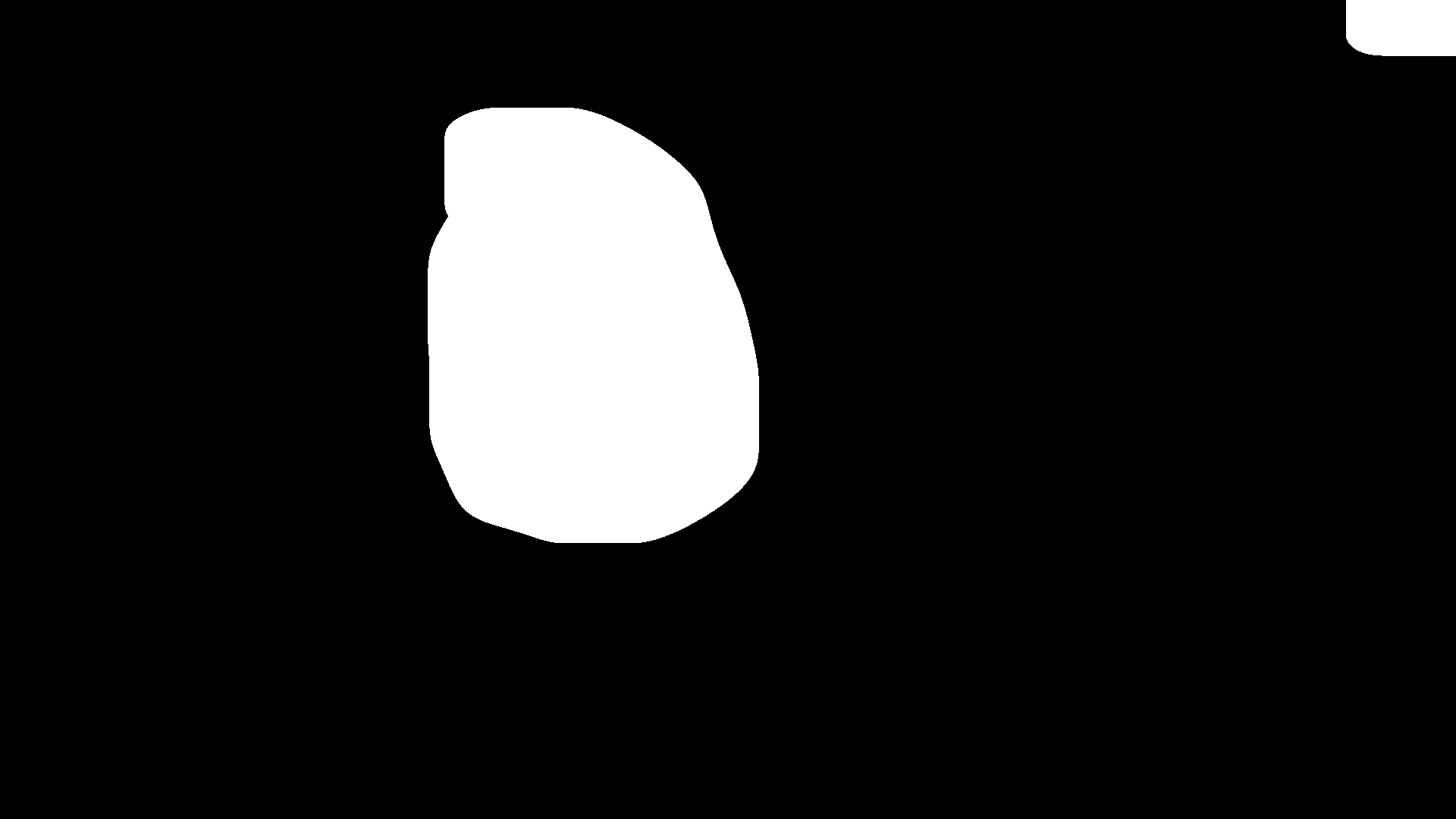}} (g) \\
    \end{minipage}
    \hfill
    \begin{minipage}[h]{0.47\linewidth}
    \center{\includegraphics[width=1\linewidth]{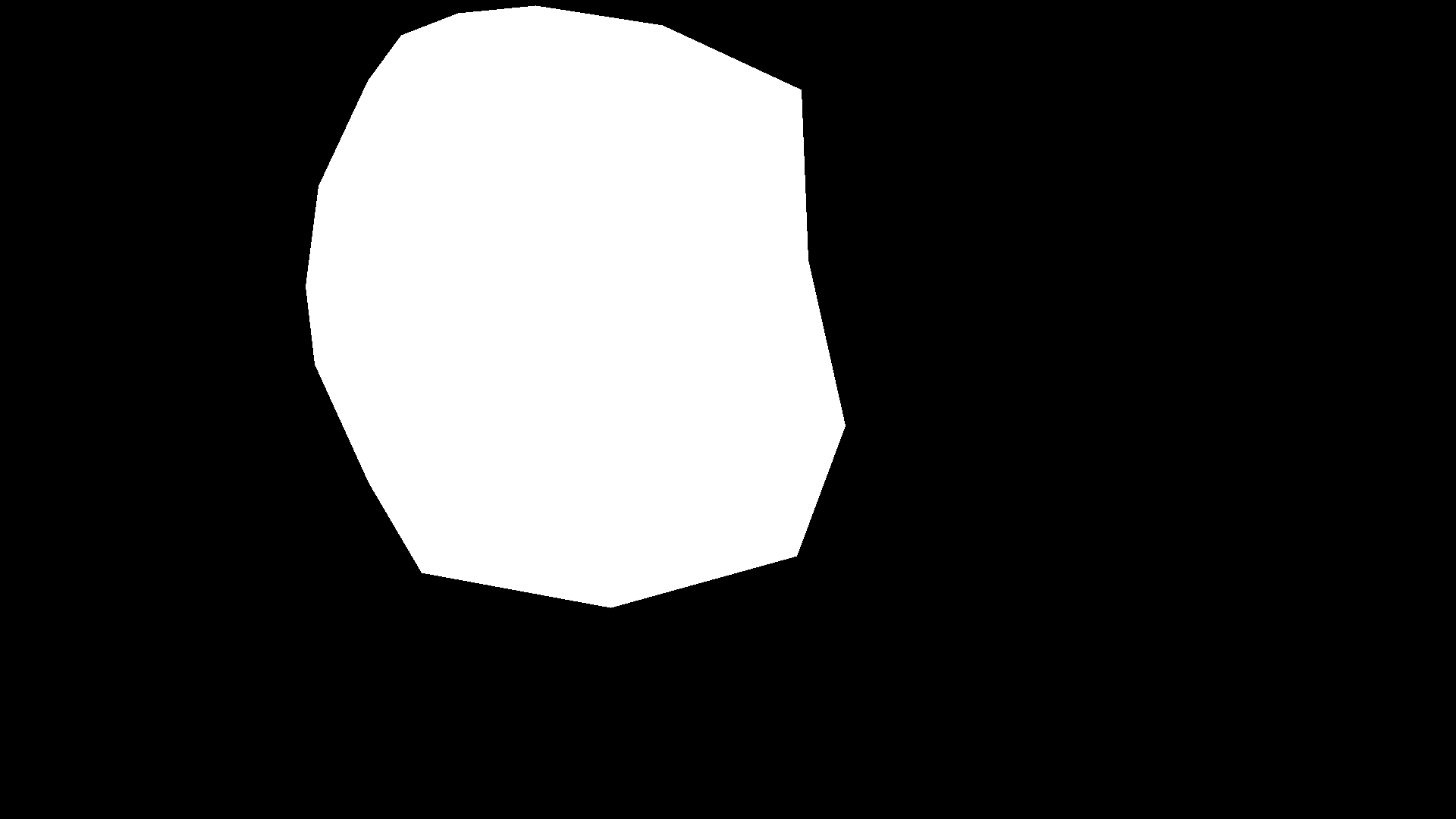}} (h) \\
    \end{minipage}
    \caption{This figure shows an example of artifact detection for a camera having an adherent raindrop near the lens center. Two images (a) and (b) are captured at $10$ frames interval.  Calculation of the averaged gradient map over these $10$ frames leads to the gradient map (c). In this map, the artifact is visible at low gradient values (dark pixels). Fig. (d) illustrates the result of averaged gradient map binarization. There is a noticeable noise present in the binarized image. To filter it, a Gaussian blur is applied to the gradient map as shown in (e) before binarization (see Fig.(f)). Finally, we get a binary segmentation mask (g) by applying dilatation to the image (f). Fig. (h) shows a ground-truth manual segmentation for this sequence.}
    \label{fig:detection}
\end{figure}

Gaussian filter is applied to the averaged gradient map to reduce salt-and-pepper noise produced by small regions without edges on an image sequence but not containing raindrops (see Fig.~\ref{fig:detection}e). For actually detecting an artifact and obtain the binary segmentation mask, the accumulated gradient map is inversely binarized (white become black and wise versa) by thresholding with a certain threshold $T_b$. Since the Gaussian filter not only reduces noise but also reduces the size of the drop itself(see Fig.~\ref{fig:detection}e as the result of applying Gaussian blur to Fig.~\ref{fig:detection}c), after binarization, dilation is applied, which returns the raindrop to its previous size. 

\begin{center}
    \begin{figure}[ht!]
    \centering
    \includegraphics[width=0.8\linewidth]{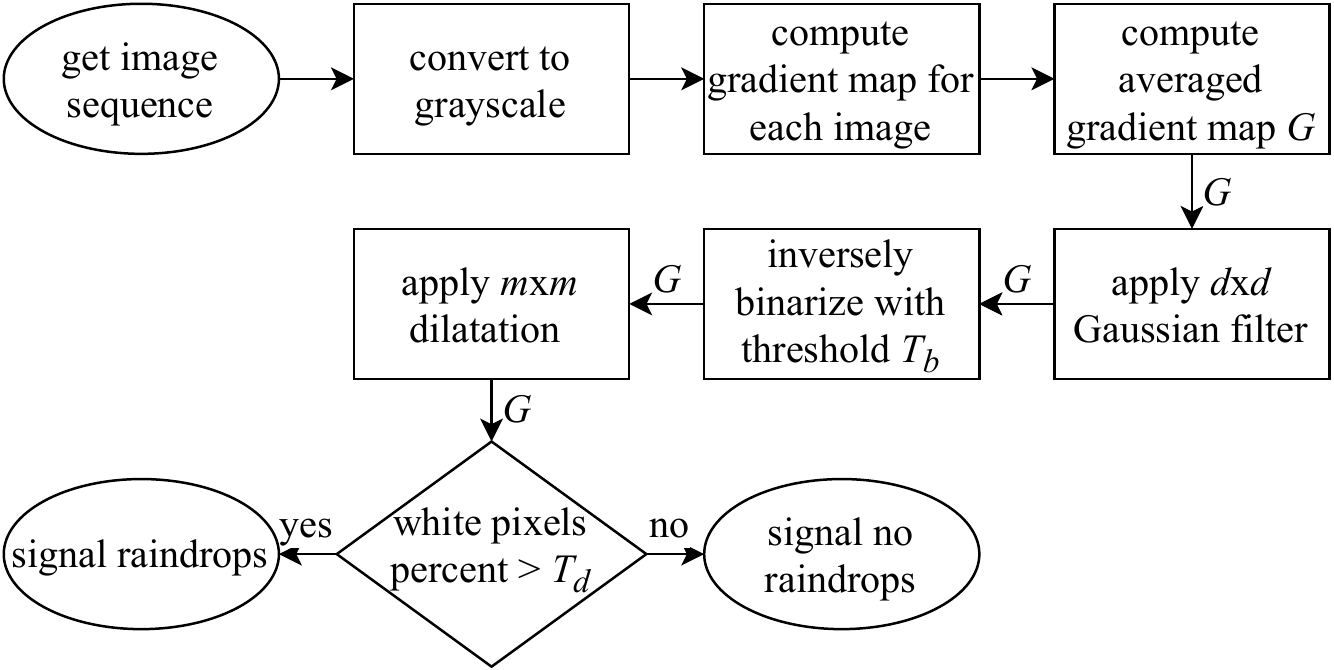}
    \caption{Raindrop detection algorithm.}
    \label{fig:algorithm}
    \end{figure}
\end{center}

For the raindrops presence detection, the percent of artifact pixels (white pixels) is calculated in the resulting binary map. If it exceeds the detection threshold $T_d$, then the algorithm signals raindrops in the sequence of images. A step-by-step raindrops detection algorithm is demonstrated in Fig.~\ref{fig:algorithm}.

\section{Experimental evaluation}
To demonstrate the applicability of the dataset for solving various problems and show the quality of the proposed algorithm, we consider the problem of detecting drops in images in two forms. The first one is the binary classification of image sequences into sequences with drops and sequences without ones. And the second one is the raindrop region segmentation.

The dataset was divided into a training set ($719$ sequences of $10$ images) and a test set ($100$ sequences of $10$ images) to select gradient-based algorithm parameters and evaluate its performance. In both training and test set, the number of sequences with drops was approximately equal to the number of sequences without ones.

For all experiments, we used a personal computer with a 6-core Intel Core i7 - 9750H, 2.6 GHz, 16.0 GB RAM.

\subsection{Algorithm parameters selection}
As described in the previous section, the gradient-based algorithm requires selecting four parameters:
\begin{itemize}
    \item $d$ the size of the Gaussian filter kernel,
    \item $T_b$ the binarization threshold,
    \item $m$ the size of the dilation kernel,
    \item $T_d$ the detection threshold.
\end{itemize}
The parameters used in all further experiments were selected based on the raindrop segmentation quality and the image processing time. The detection threshold -- the area occupied by raindrops on the image to signal the successful detection -- was fixed as $T_d = 0.1$. The choice of such a value for the detection threshold is based on the fact that overlapping only ten percent of the image can make the autonomous system further movement unsafe.

The area occupied by drops is small relative to the image size, which makes segmentation classes unbalanced.  For all standard metrics such as Intersection over Union (IoU) or Dice coefficient, the highest values are obtained when the algorithm signals the total absence of raindrops.

To solve this, we used the Dice coefficient accumulated over all sequences as follows: if there are $L$ sequences and $I_i$ is the intersection of the ground truth raindrop area and the binary mask found by the algorithm for sequence number $i$, and $U_i$ is the sum of pixels belonging to the raindrop area in both images, then the metric is computed as follows:
$$\text{Accumulated-Dice}(d,m,T_b,T_d) = \dfrac{\sum\limits_{i=1}^L I_i}{ \sum\limits_{i=1}^L U_i}. $$
To select the optimal parameters, we tried to maximize this accumulated Dice coefficient and minimize image processing time. With this metrics we obtained the following optimal parameter values: $d = 271$, $m = 91$, $T_b = 0.18$.

\subsection{Image sequence classification}

The problem of detecting raindrops on a sequence of images was considered as a binary classification of image sequences. The quality of the sequence classification was assessed using ROC curves and the AUC-ROC metric. ROC curves were generated by varying the free parameter $T_b$ -- the threshold for gradient map binarization. In all experiments, $T_b$ varied from $0.1$ to $0.9$ with a step of $0.05$. 

\begin{figure}[t]
    \begin{minipage}[h]{0.49\linewidth}
    \center{\includegraphics[width=1\linewidth]{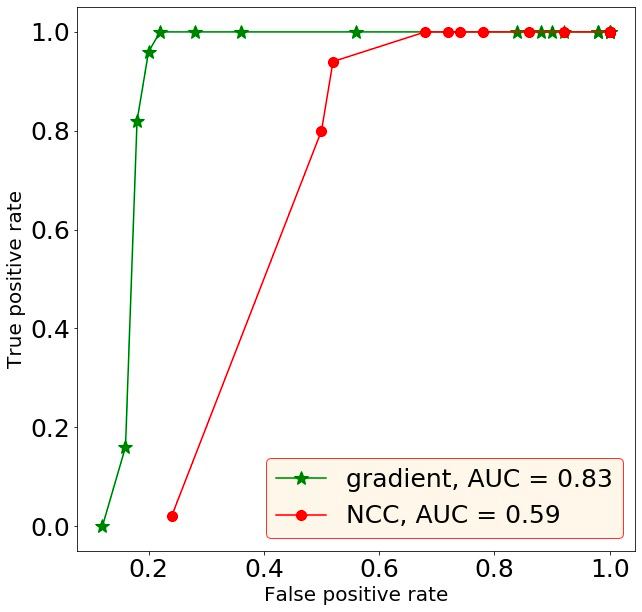}} (a) \\
    \end{minipage}
    \hfill
    \begin{minipage}[h]{0.49\linewidth}
    \center{\includegraphics[width=1\linewidth]{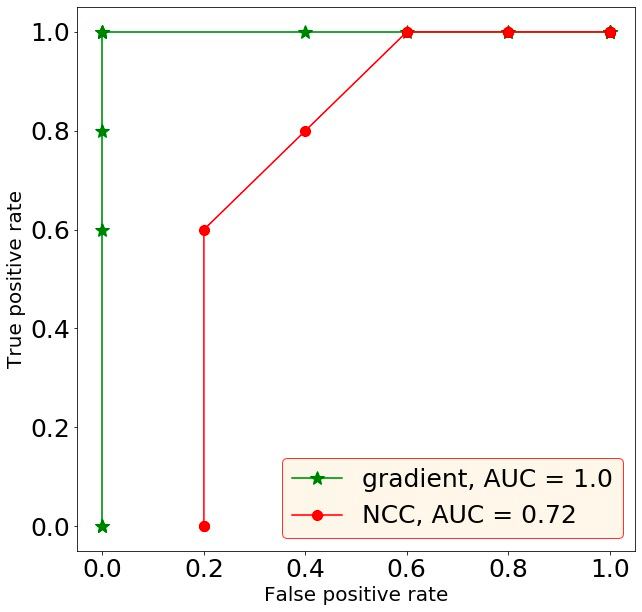}} \\(b)
    \end{minipage}
\begin{center}
\includegraphics[width = 0.49\linewidth]{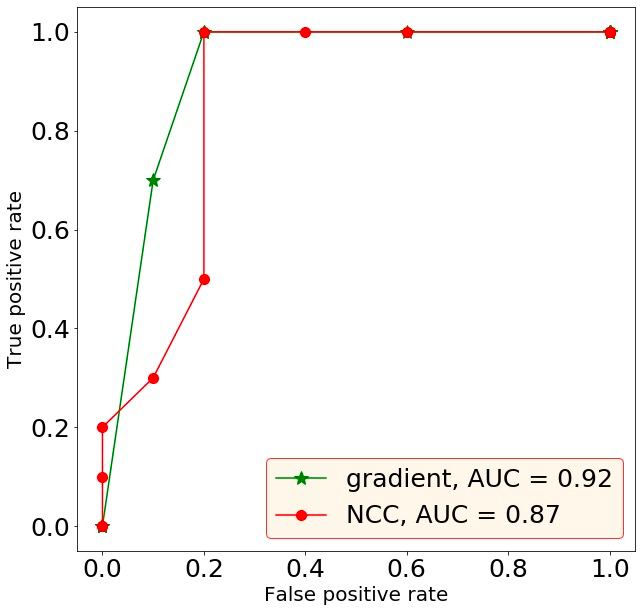} \\(c)
\caption{ROC-curves of the proposed algorithm and the algorithm based on normalized cross-correlation (NCC) from \cite{Einecke2014}. ROC-curves plotted over: (a) -  sequences consisting of $10$ images, (b) - sequences consisting of $100$ images, (c) - sequences consisting of $10$ images and taken during the turning maneuver.}
\label{fig:roc round}

\end{center}
\end{figure}

Fig.~\ref{fig:roc round}(a) shows the ROC curve of the gradient raindrop detection algorithm calculated on sequences of $10$ images, using the $5 \times 5$ Sobel operator. The value of the AUC-ROC metric in this experiment turned out to be $ 0.83 $, and the processing time for the whole sequence varies from $ 0.7 $ to $ 0.8 $ seconds. Thus, the algorithm provides a high-quality classification of sequences with a short image processing time, making it potentially applicable for autonomous visual systems self-check procedure.

We also tested our algorithm on sequences of $100$ images. For this, $100$ sequences of $10$ images were re-sorted into $10$ sequences of $100$ images, which was possible since together all these images made up a complete video. As expected, the results showed that an increase in the number of images in sequences leads to increased classification quality. Although it also increases the time required for the detection of raindrops by $10$ times. The ROC-AUC metric in this experiment turned out to be $1.0$. However, since the sample consisted of only $10$ sequences, this does not mean that the algorithm works perfectly. 

We compared the proposed algorithm with the existing one based on pixel-wise normalized cross-correlation (NCC)~\cite{Einecke2014}. According to the authors, the NCC's main drawback is its poor performance on sequences taken during a straight motion. In this case, distant objects in the images become almost stationary relative to the observer, which leads to high correlation values between subsequent frames and false-positive raindrop detections. Therefore, to compare our algorithm with NCC, among all sequences of $10$ images, we selected only those that were captured during the turning maneuver. So, we got $20$ sequences, among which a half were with raindrops.

The ROC-curves received as a result of testing two algorithms on such sequences are shown in~Fig.~\ref{fig:roc round}c. As it could be seen from the plots, both algorithms demonstrate a high quality of classification. However, the gradient-based algorithm shows the best metric value: AUC-ROC(Grad) = $0.92$, AUC-ROC(NCC) = $0.87$. In addition, ROC-curves on a full set of sequences (~Fig.~\ref{fig:roc round}a-b) showed that in contrast to NCC, the proposed algorithm also works well on sequences captured during linear motion. This property gives the gradient-based algorithm a great advantage since autonomous vehicle movement is mainly rectilinear.

Also, we compared the proposed algorithm and NCC in image processing time. Although in the implementation with the box filter~\cite{mcdonnell1980}, NCC, like the proposed algorithm, has a linear computational complexity of one image processing, it turned out that the processing time for one sequence of $10$ images for NCC is $1.4-1.5$ seconds, which is almost two times longer than the processing time for one sequence by the proposed algorithm, which is $0.7-0.8$ seconds.

\subsection{Raindrop region segmentation}
To demonstrate the dataset applicability for solving the raindrop segmentation problem, we trained a neural network of the original U-Net architecture~\cite{UNet2015}. As a training sample, we took the same set of $7190$ images on which the gradient-based algorithm parameters were selected. Also, $500$ images with artificial drops were added to the training set to augment the data. The test set consisted of the same $1000$ images on which the gradient-based method was tested in the previous section. All the images in this experiment were resized to $216\times216$.

Three metrics were used to assess the quality of segmentation: Dice coefficient calculated over raindrops pixels, Intersection over Union, and accuracy. The average values of all three metrics on the test sample turned out to be very high: IoU = $0.86$, Dice = $0.82$, accuracy = $0.97$. It means that the dataset suits well for solving the segmentation problem and training neural networks.

Besides, the proposed gradient-based algorithm was also evaluated on solving segmentation problem on a sequence of images. In this case, all steps of the algorithm remain the same until obtaining a binary mask. Metric values for the gradient in this experiment on sequences of $10$ images resized to $216\times216$ were as follows: IoU = $0.72$, Dice = $0.5$, accuracy = $0.95$. As expected, U-Net significantly outperforms the proposed method. On the other hand, the gradient-based algorithm requires significantly less time to process one image. The average processing time for one image for U-Net is $0.7$ second, while the same time for the proposed algorithm is $0.025$ seconds. Thus, the gradient method loses in segmentation quality compared to the U-Net. However, it can work in real-time, consuming less computational resources, which is very important for autonomous vehicles. Detailed results of experiments are presented in table~\ref{tab:experiments}.

\begin{table}[!t]
    
\centering
\caption{Results of comparing quality and speed of raindrop segmentation for the gradient-based algorithm and U-Net. Dice score calculated over raindrops pixels, Intersection over Union, pixel-wise accuracy and average one $216\times216$ image processing time were used to evaluate the performances of the algorithm.} 
    \begin{tabular}{|c|c|c|c|c|}
    \hline
    \multicolumn{1}{|c|}{}   & IoU  & Dice & Accuracy & Time(s)  \\ \hline
    Gradient-based algorithm & 0.72 & 0.50 & 0.95     & 0.025 \\ \hline
    UNet                     & 0.86 & 0.82 & 0.97     & 0.7  \\ \hline
    \end{tabular}
    \label{tab:experiments}
\end{table}

\section{Conclusion}

In this paper, we presented a publicly available dataset for training and evaluating algorithms for various problems of detecting raindrops on the camera lens or windshield in front of the camera. The dataset contains sequences of images captured by a camera attached to the vehicle during its movement. At the moment, it consists of $8190$ images, 3390 of which contain raindrops. All images are annotated with the binary mask representing areas with raindrops.

We demonstrated the applicability of the dataset in raindrop presence detection and raindrop segmentation problems. In addition to the dataset, we presented an artificial raindrops generation algorithm, which allows the creation of various realistic raindrops to augment the data. 

At the moment, we are actively working on expanding the dataset. One of the main directions of dataset development is collecting images with other weather-related artifacts such as glare, dirt, snow, and others.

Apart from the dataset, we proposed a novel algorithm for detecting raindrops on a camera lens from a sequence of images. The problem of detecting raindrops was considered as a problem of binary classification of image sequences. The performance of the algorithm was evaluated using ROC curves and AUC-ROC metrics. The experiment results showed that the algorithm reliably detects raindrops and demonstrates better classification quality and higher image processing speed than the already existing algorithm~\cite{Einecke2014}. 

Also, the gradient-based algorithm was tested in the problem of raindrop segmentation. We trained the original U-Net architecture's neural network~\cite{UNet2015} to compare with our algorithm and demonstrate the dataset applicability for this task. To evaluate the algorithms' segmentation quality, we used Dice score calculated over raindrops pixels, Intersection over Union, and pixel-wise accuracy. According to the experiment results, U-Net showed higher metrics values on the test sample than the gradient-based algorithm. However, the proposed algorithm requires significantly less time to process one image. Thus, it can work in real-time, consuming less computational resources, which is essential for an autonomous vehicle.

Further improvement of the proposed algorithm can be aimed at developing additional verification of the detected raindrops to avoid misinterpretation by the algorithm of other homogeneous and static regions, such as the sky. Also, this algorithm can potentially be used to detect other types of artifacts besides raindrops, such as snow, dirt, scratches on the lens, and others.

\bibliographystyle{unsrt}
\bibliography{mybib}

\end{document}